\documentclass[11pt,a4paper]{article}
\usepackage[hyperref]{acl2017}
\usepackage{times}
\usepackage{latexsym}

\usepackage{url}

\usepackage{arydshln}

\usepackage{amsmath}
\usepackage{url}
\usepackage{stmaryrd}
\usepackage{qtree}
\usepackage{tikz-qtree}
\usepackage{graphicx}
\usetikzlibrary{matrix}
\usetikzlibrary{chains}
\usetikzlibrary{decorations.pathreplacing,shapes,backgrounds}
\usetikzlibrary{fit}
\usepackage{array}
\usepackage{alltt}
\usepackage{listings}

\usepackage{CJK,CJKspace,CJKpunct}

\usepackage{array}
\usepackage{tabulary}
\newcolumntype{K}[1]{>{\centering\arraybackslash}p{#1}}

\usepackage{algorithm}
\usepackage[noend]{algpseudocode}
\usepackage{pifont}
\usepackage{etoolbox}\AtBeginEnvironment{algorithmic}{\footnotesize}

\usepackage{color, colortbl}
\definecolor{Gray}{gray}{0.9}
\definecolor{darkolivegreen}{rgb}{0.33, 0.42, 0.18}
\definecolor{darkseagreen}{rgb}{0.56, 0.74, 0.56}
\usepackage{amssymb}

\aclfinalcopy 

\newcolumntype{P}[1]{>{\raggedright\arraybackslash}p{#1}}

\title{Learning Semantic Correspondences in Technical Documentation}

\author{Kyle Richardson \and Jonas Kuhn\\
   Institute of Natural Language Processing  \\
  University of Stuttgart \\
  {\tt \{kyle,jonas\}@ims.uni-stuttgart.de} 
  \\}

\date{}

\begin{document}
\maketitle
\begin{abstract}
We consider the problem of translating high-level textual descriptions to formal representations in technical documentation as part of an effort to model the meaning of such documentation.  We focus specifically on the problem of learning translational correspondences between text descriptions and grounded representations in the target documentation, such as formal representation of functions or code templates.  Our approach exploits the parallel nature of such documentation, or the tight coupling between high-level text and the low-level representations we aim to learn. Data is collected by mining technical documents for such parallel text-representation pairs, which we use to train a simple semantic parsing model. We report new baseline results on sixteen novel datasets, including the standard library documentation for nine popular programming languages across seven natural languages, and a small collection of Unix utility manuals. 




\end{abstract}

\section{Introduction}


Technical documentation in the computer domain, such as source code documentation and other how-to manuals, provide high-level descriptions of how lower-level computer programs and utilities work. Often these descriptions are coupled with formal representations of these lower-level features, expressed in the target programming languages. For example, Figure 1.1 shows the source code documentation (in red/bold) for the \texttt{max} function in the Java programming language paired with the representation of this function in the underlying Java language (in black). This formal representation captures the name of the function, the return value, the types of arguments the function takes, among other details related to the function's place and visibility in the overall source code collection or API.


\begin{figure}[t]
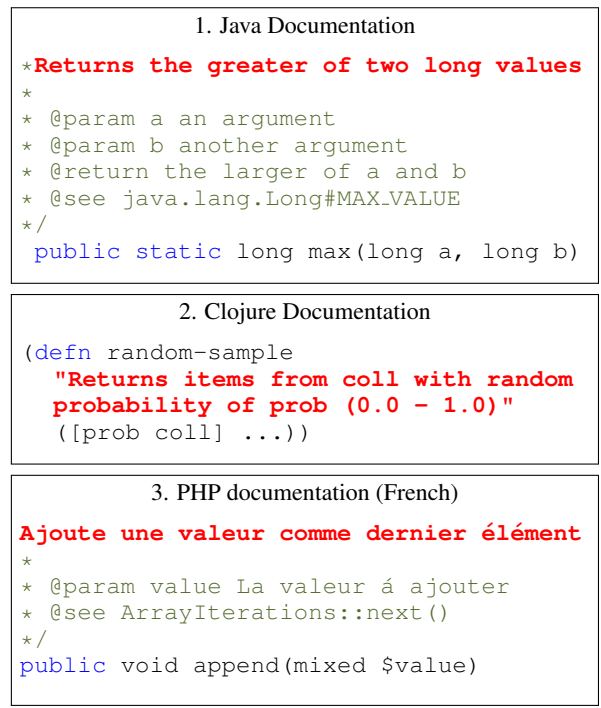

\noindent\fbox{%
    \parbox{7.5cm}{%
\begin{centering}    
{\footnotesize \text{1. Java Documentation}}  \\[-.7cm]    
\end{centering}
     {\footnotesize \begin{alltt}\textcolor{darkolivegreen}{
%/* \\     
*\textcolor{red}{\textbf{Returns the greater of two long values}} \\
* \\
* \text{@param} a an argument \\
* \text{@param} b another argument \\
* \text{@return} the larger of a and b \\
* \text{@see} java.lang.Long\#MAX\_VALUE \\
*/ \\ }
\text{\textcolor{blue}{public static} long max(long a, long b)}
    \end{alltt}
    }}}
\\[.1cm]
\noindent\fbox{%
    \parbox{7.5cm}{
\begin{centering}    
{\footnotesize \text{2. Clojure Documentation}}  \\[-.7cm]       
\end{centering}
     {\footnotesize \begin{alltt}
%     \textcolor{darkolivegreen}{
%\text{(ns ... clojure.core)} \\[-.2cm] \\}
(\textcolor{blue}{\text{defn}} \text{random-sample}\\
\hspace*{.1in} \textcolor{red}{\textbf{"Returns items from coll with random }} \\
\hspace*{.1in} \textcolor{red}{\textbf{probability of prob (0.0 - 1.0)" }} \\
%\hspace*{.1in} ([\text{prob}] ...)  \\
\hspace*{.1in} ([\text{prob coll}] ...)) 
    \end{alltt}
    }}}
\\[.1cm]
\noindent\fbox{%
    \parbox{7.5cm}{%
\begin{centering}
{\footnotesize \text{3. PHP documentation (French)}}  \\[-.7cm]      
\end{centering}    
     {\footnotesize \begin{alltt}\textcolor{darkolivegreen}{
%namespace ArrayIterator; \\[.3cm]     
%/* \\
\textcolor{red}{\textbf{Ajoute une valeur comme dernier \'el\'ement}} \\
* \\
* \text{@param} value La valeur \'a ajouter \\
* \text{@see} ArrayIterations::next()  \\
*/ \\ }\text{\textcolor{blue}{public} void  append(mixed \$value)} \vspace{.1cm} \end{alltt}}}}

\caption{Example source code documentation. }
\end{figure}

Given the high-level nature of the textual annotations, modeling the meaning of any given description is not an easy task, as it involves much more information than what is directly provided in the associated documentation.  For example, capturing the meaning of the description \emph{the greater of} might require having a background theory about quantity/numbers and relations between different quantities. A first step towards capturing the meaning, however, is learning to translate this description to symbols in the target representation, in this case to the \texttt{max} symbol. By doing this translation to a formal language, modeling and learning the subsequent semantics becomes easier since we are eliminating the ambiguity of ordinary language. Similarly, we would want to first translate the description \emph{two long values}, which specifies the number and type of argument taken by this function, to the sequence \texttt{long a,long b}.


\begin{figure}[t]
\noindent\fbox{%
    \parbox{7.5cm}{%
\begin{centering}
{\footnotesize \text{Unix Utility Manual}}  \\[-.5cm]      
\end{centering}    
{\footnotesize \begin{alltt}
\text{\textcolor{blue}{NAME}} : dappprof \\[.1cm]
	\hspace*{.1in} profile  user and lib function usage. \\ \\[-.2cm]
\text{\textcolor{blue}{SYNOPSIS} dappprof [-ac] -p PID | command} \\[.1cm]
	%\hspace*{.1in} dappprof [-ac..] .. {-p PID | command } \\ \\[-.2cm]
\text{\textcolor{blue}{DESCRIPTION}} \\[.1cm]
	%\hspace*{.1in} \textbf{--a} \hspace*{.4in} print all data \\
	\hspace*{.1in} \textbf{-p PID} \hspace*{.1in}  examine the PID ... \\ \\[-.2cm]
\text{\textcolor{blue}{EXAMPLES}} \\ \\[-.2cm]
	% \hspace*{.05in} \textcolor{red}{\textbf{Run and examine the ``df -h'' command}} \\[.1cm]
	%\hspace*{.3in} \textbf{dappprof} \emph{command}=\textbf{``df -h''} \\ \\
	\hspace*{.05in} \textcolor{red}{\textbf{Print elapsed time for PID 1871}} \\[.1cm]
	\hspace*{.3in}  \textbf{dappprof -p} \emph{PID}=\textbf{1871} \\ \\[-.2cm]
%	\hspace*{.01in} \textcolor{red}{\textbf{Print all data for PID 1871}} \\
%	\hspace*{.1in}  dapprof -ap \textbf{PID}=1871 \\ \\[-.1cm]
\text{\textcolor{blue}{SEE ALSO}: dapptrace(1M), dtrace(1M), ...} 	\\ [-.4cm]
	%\hspace*{.1in} dapptrace(1M), dtrace(1M), ...
\end{alltt}}        
            }
}
\caption{An example computer utility manual in the Unix domain. Descriptions of example uses are shown in red. }
\label{fig:docs}
\end{figure}

By focusing on translation, we can create new datasets by mining these types of source code collections for sets of parallel text-representation pairs. Given the wide variety of available programming languages, many such datasets can be constructed, each offering new challenges related to differences in the formal representations used by different programming languages. Figure 1.2 shows example documentation for the Clojure programming language, which is part of the Lisp family of languages. In this case, the description \emph{Returns random probability of} should be translated to the function name \texttt{random-sample} since it describes what the overall function does. Similarly,  the argument descriptions \emph{from coll} and \emph{of prob} should translate to \texttt{coll} and \texttt{prob}. 



Given the large community of programmers around the world, many source code collections are available in languages other than English. Figure 1.3 shows an example entry from the French version of the PHP standard library, which was translated by volunteer developers. Having multi-lingual data raises new challenges, and broadens the scope of investigations into this type of semantic translation. 


Other types of technical documentation, such as utility manuals, exhibit similar features. Figure 2 shows an example manual in the domain of Unix utilities. The textual description in red/bold describes an example use of the \emph{dappprof} utility paired with formal representations in the form of executable code. As with the previous examples, such formal representations do not capture the full meaning of the different descriptions, but serve as a convenient operationalization, or \emph{translational semantics}, of the meaning in Unix. \emph{Print elapsed time}, for example, roughly describes what the \texttt{dappprof} utility does, whereas \emph{PID 1871} describes the second half of the code sequence. 


In both types of technical documentation, information is not limited to raw pairs of descriptions and representations, but can include other information and clues that are useful for learning. Java function annotations include textual descriptions of individual arguments and return values (shown in green). Taxonomic information and pointers to related functions or utilities are also annotated (e.g., the \texttt{@see} section in Figure 1, or \texttt{SEE ALSO} section in Figure 2). Structural information about code sequences, and the types of abstract arguments these sequences take, are described in the \texttt{SYNOPSIS} section of the Unix manual. This last piece of information allows us to generate abstract code templates, and generalize individual arguments. For example, the raw argument \texttt{1871} in the sequence \texttt{dappprof -p 1871} can be typed as a \texttt{PID} instance, and an argument of the \texttt{-p} flag.   



Given this type of data, a natural experiment is to see whether we can build programs that translate high-level textual descriptions to correct formal representations. We  aim to learn these translations using raw text-meaning pairs as the sole supervision. Our focus is on learning function translations or representations within nine programming language APIs, each varying in size, representation style, and source natural language. To our knowledge, our work is the first to look at translating source code descriptions to formal representations using such a wide variety of programming and natural languages. In total, we introduce fourteen new datasets in the source code domain that include seven natural languages, and report new results for an existing dataset.  As well, we look at learning simple code templates using a small collection of English Unix manuals. 

The main goal of this paper is to establish strong baselines results on these resources, which we hope can be used for benchmarking and developing new semantic parsing methods. We achieved initial baselines using the language modeling and translation approach of \newcite{deng}. We also show that modest improvements can be achieved by using a more conventional discriminative model \cite{ZettlemoyerO} that, in part, exploits document-level features from the technical documentation sets. 







\section{Related Work}

Our work is situated within research on semantic parsing, which focuses on the problem of generating formal meaning representations from text for natural language understanding applications.  Recent interest in this topic has centered around learning meaning representation from example text-meaning pairs, for applications such as automated question-answering \cite{berant2013semantic}, robot control \cite{ROBOT} and text generation \cite{wong2007generation}. 


While generating representations for natural language understanding is a complex task, most studies focus on the translation or generation problem independently of other semantic or knowledge representation issues. Earlier work looks at supervised learning of logical representations using example text-meaning pairs using tools from statistical machine translation \cite{wongMAIN} and parsing \cite{ZettlemoyerO}. These methods are meant to be applicable to a wide range of translation problems and representation types, which make new parallel datasets or resources useful for furthering the research. 

In general, however, such datasets are hard to construct since building them requires considerable domain knowledge and knowledge of logic. Alternatively, we construct parallel datasets automatically from technical documentation, which obviates the need for annotation. While the formal representations are not actual logical forms, they still provide a good test case for testing how well semantic parsers learn translations to representations.

To date, most benchmark datasets are limited to small controlled domains, such as geography and navigation. While attempts have been made to do open-domain semantic parsing using larger, more complex datasets {\cite{berant2013semantic,pasupatcompositional}, such resources are still scarce. In Figure 3, we compare the details of one widely used dataset, Geoquery \cite{zelle}, to our new datasets. Our new resources are on average much larger than geoquery in terms of the number of example pairs, and the size of the different language vocabularies. Most existing datasets are also primarily English-based, while we focus on learning in a multilingual setting using several new moderately sized datasets. 

Within semantic parsing, there has also been work on situated or grounded learning, that involves learning in domains with weak supervision and indirect cues \cite{liang2016learning,richardson2016learning}. This has sometimes involved learning from automatically generated parallel data and representations \cite{chenMain} of the type we consider in this paper. Here one can find work in technical domains, including learning to generate regular expressions \cite{manshadi2013integrating,kushman2013using} and other types of source code \cite{quirk2015language}, which ultimately aim to solve the problem of natural language programming. We view our work as one small step in this general direction. 

Our work is also related to software components retrieval and builds on the approach of \newcite{deng}.  Robustly learning the translation from language to code representations can help to facilitate natural language querying of API collections \cite{lv2015codehow}.   As part of this effort, recent work in machine learning has focused on the similar problem of learning code representations using resources such as StackOverflow and Github. These studies primarily focus on learning longer programs \cite{allamanis2015bimodal} as opposed to function representations, or focus narrowly on a single programming language such as Java \cite{gu2016deep} or on related tasks such as text generation \cite{code_generate,oda2015learning}. To our knowledge, none of this work has been applied to languages other than English or such a wide variety of programming languages.



\begin{figure*}[t]
\centering

\newcommand{\PreserveBackslash}[1]{\let\temp=\\#1\let\\=\temp}
\newcolumntype{L}[1]{>{\PreserveBackslash\raggedright}p{#1}}
\newcolumntype{P}[1]{>{\raggedright\arraybackslash}p{#1}}

\begin{tabular}{| P{1.1cm} | P{.7cm} P{.7cm} P{.7cm} P{.7cm} l | P{8.3cm} |}
\hline
{\footnotesize Dataset} & {\footnotesize \#Pairs} & {\footnotesize \#Descr.} & {\footnotesize Symbols} & {\footnotesize \#Words} & {\footnotesize Vocab.} & {\footnotesize Example Pairs $(x,z)$, \textbf{Goal: }learn a function $x \to z$}\tabularnewline \hline
{\footnotesize \text{Java}}  & {\footnotesize \text{7,183}} & {\footnotesize 4,804} & {\footnotesize 4,072} & {\footnotesize 82,696}  & {\footnotesize 3,721}  & 
\begin{tabular}{p{.1cm}l}
{\footnotesize $x:$ } & {\footnotesize Compares this Calendar to the specified Object.} \\[-.1cm]
{\footnotesize $z:$ } & {\footnotesize \texttt{boolean util.Calendar.equals(Object obj)}} \tabularnewline
\end{tabular} \\ \hline
{\footnotesize \text{Ruby}}  & {\footnotesize \text{6,885}} & {\footnotesize 1,849} & {\footnotesize 3,803} & {\footnotesize 67,274}  & {\footnotesize 5,131}  & 
\begin{tabular}{p{.1cm}l}
{\footnotesize $x:$ } & {\footnotesize Computes the arc tangent given y and x.} \\[-.1cm]
{\footnotesize $z:$ } & {\footnotesize \texttt{Math.atan2(y,x) $\to$ Float}} \tabularnewline
\end{tabular} \\ \hline
{\footnotesize {\text{PHP$_{en}$}}} & {\footnotesize \text{6,611}}  &  {\footnotesize 13,943} &  {\footnotesize 8,308} & {\footnotesize 68,921} &  {\footnotesize 4,874}  &
\begin{tabular}{p{.1cm}l}
{\footnotesize $x:$ } & {\footnotesize Delete an entry in the archive using its name.}  \\[-.1cm]
{\footnotesize $z:$ } & {\footnotesize \texttt{bool ZipArchive::deleteName(string \$name)}} \tabularnewline
\end{tabular}\tabularnewline \hline
{\footnotesize {\text{Python}}} & {\footnotesize \text{3,085}}  &  {\footnotesize 429} &  {\footnotesize 3,991} & {\footnotesize 27,012} &  {\footnotesize 2,768}  & 
\begin{tabular}{p{.1cm}l}
{\footnotesize $x:$ } & {\footnotesize Remove the specific filter from this handler.}  \\[-.1cm]
{\footnotesize $z:$ } & {\footnotesize \texttt{logging.Filterer.removeFilter(filter)}} \tabularnewline
\end{tabular}\tabularnewline \hline
{\footnotesize {\text{Elisp}}} & {\footnotesize \text{2,089}}  &  {\footnotesize 1,365} &  {\footnotesize 1,883} & {\footnotesize 30,248} &  {\footnotesize 2,644}  & 
\begin{tabular}{p{.1cm}l}
{\footnotesize $x:$ } & {\footnotesize This function returns the total height, in lines,  of the window.}  \\[-.1cm]
{\footnotesize $z:$ } & {\footnotesize \texttt{(window-total-height window round)}} \tabularnewline
\end{tabular}\tabularnewline \hline
{\footnotesize {\text{Haskell}}} & {\footnotesize \text{1,633}} & {\footnotesize 255} & {\footnotesize 1,604} & {\footnotesize 19,242} & {\footnotesize 2,192} & %
\begin{tabular}{p{.1cm}l}
{\footnotesize $x:$ } & {\footnotesize Extract the second component of a pair.} \\[-.1cm]
{\footnotesize $z:$ } & {\footnotesize \texttt{Data.Tuple.snd :: (a, b) -> b}}\tabularnewline
\end{tabular}\tabularnewline \hline
{\footnotesize {\text{Clojure}}} & {\footnotesize \text{1,739}}  &  {\footnotesize --} &  {\footnotesize 2,569} & {\footnotesize 17,568} &  {\footnotesize 2,233}  & %
\begin{tabular}{p{.1cm}l}
{\footnotesize $x:$ } & {\footnotesize Returns a lazy seq of every nth item in coll.}  \\[-.1cm]
{\footnotesize $z:$ } & {\footnotesize \texttt{(core.take-nth n coll)}} \tabularnewline
\end{tabular}\tabularnewline \hline
{\footnotesize {\text{C}}} & {\footnotesize \text{1,436}}  &  {\footnotesize  1,478} &  {\footnotesize 1,452} & {\footnotesize 12,811} &  {\footnotesize 1,835}  & %
\begin{tabular}{p{.1cm}l}
{\footnotesize $x:$ } & {\footnotesize Returns the current file position of the stream stream.}  \\[-.1cm]
{\footnotesize $z:$ } & {\footnotesize \texttt{long int ftell(FILE *stream)}} \tabularnewline
\end{tabular}\tabularnewline \hline
{\footnotesize {\text{Scheme}}} & {\footnotesize \text{1,301}}  &  {\footnotesize 376} &  {\footnotesize 1,343} & {\footnotesize 15,574} &  {\footnotesize 1,756}  & %
\begin{tabular}{p{.1cm}l}
{\footnotesize $x:$ } & {\footnotesize Returns a new port with type port-type and the given state.}  \\[-.1cm]
{\footnotesize $z:$ } & {\footnotesize \texttt{(make-port port-type state)}} \tabularnewline
\end{tabular}\tabularnewline 
 \hline\hline
 {\footnotesize {\text{Unix}}} & {\footnotesize \text{921}}  &  {\footnotesize 940} &  {\footnotesize 1,000} & {\footnotesize 11,100} &  {\footnotesize 2,025}  & %
\begin{tabular}{p{.1cm}l}
{\footnotesize $x:$ } & {\footnotesize To get policies for a specific user account.}  \\[-.1cm]
{\footnotesize $z:$ } & {\footnotesize \texttt{pwpolicy -u username -getpolicy}} \tabularnewline
\end{tabular}\tabularnewline 
 \hline\hline
{\footnotesize\text{Geoquery}} & {\footnotesize \text{880}} & {\footnotesize --} & {\footnotesize {167}} & {\footnotesize 6,663} & {\footnotesize {279}} & 
\begin{tabular}{p{-.7cm}l}
{\footnotesize $x:$ } & {\footnotesize What is the tallest mountain in America?}  \\[-.1cm]
{\footnotesize $z:$ } & {\footnotesize \texttt{(highest(mountain(loc\_2(countryid usa))))}} \tabularnewline
\end{tabular}\tabularnewline \hline

\end{tabular}

\caption{Description of our  English corpus collection with example text/function pairs.}
\end{figure*}

\section{Mapping Text to Representations}

In this section, we formulate the basic problem of translating to representations in technical documentation.

\subsection{Problem Description}

We use the term \emph{technical documentation} to refer to two types of resources: textual descriptions inside of source code collections, and computer utility manuals. In this paper, the first type includes high-level descriptions of functions in standard library source code documentation.  The second type includes a collection of Unix manuals, also known as man pages. Both types include pairs of text and code representations.

We will refer to the target representations in these resources as \emph{API components}, or components. In source code, components are formal representations of functions, or \emph{function signatures} \cite{deng}. The form of a function signature varies depending on the resource, but in general gives a specification of how a function is named and structured. The example function signatures in Figure 3 all specify a function name, a list of arguments, and other optional information such as a return value and a namespace. Components in utility manuals are short executable code sequences intended to show an example use of a utility. We assume typed code sequences following \newcite{unixman}, where the constituent parts of the sequences are abstracted by type. 


Given a set of example text-component pairs, $D = \{(x_{i},z_{i})\}_{i=1}^{n}$, the goal is to learn how to generate correct, well-formed components $z \in \mathcal{C}$ for each input $x$. Viewed as a semantic parsing problem, this treats the target components as a kind of formal meaning representation, analogous to a logical form. In our experiments, we assume that the complete set of output components are known. In the API documentation sets, this is because each standard library contains a finite number of function representations, roughly corresponding to the number of pairs as shown in Figure 3. For a given input, therefore, the goal is to find the best candidate function translation within the space of the total API components $\mathcal{C}$ \cite{deng}. 


Given these constraints, our setup closely resembles that of \newcite{kushman}, who learn to parse algebra word problems using a small set of equation templates. Their approach is inspired by template-based information extraction, where templates are recognized and instantiated by slot-filling. Our function signatures and code templates have a similar slot-like structure, consisting of slots such as return value, arguments, function name and namespace. 

\subsection{Language Modeling Baselines} 

Existing approaches to semantic parsing formalize the mapping from language to logic using a variety of formalisms including CFGs \cite{BB}, CCGs \cite{Kwiatkowski}, synchronous CFGs \cite{wongLAM}. Deciding to use one formalism over another is often motivated by the complexities of the target representations being learned. For example, recent interest in learning graph-based representations such as those in the AMR bank \cite{banarescu2013abstract} requires parsing models that can generate complex graph shaped derivations such as CCGs \cite{artzi2015broad} or HRGs \cite{peng2015synchronous}. Given the simplicity of our API representations, we opt for a simple semantic parsing model that exploits the finiteness of our target representations. 




Following (\cite{deng}; henceforth DC), we treat the problem of component translation as a language modeling problem \cite{languagemodeling}. For a given \emph{query} sequence or text $x = w_{i},..,w_{I}$ and \emph{component} sequence $z = u_{j},..,u_{J}$, the probability of the component given the query is defined as follows using Bayes' theorem: $p(z \vert x) \propto p(x \vert z) p(z)$. 


By assuming a uniform prior over the probability of each component $p(z)$, the problem reduces to computing $p(x \vert z)$, which is where language modeling is used. Given each word $w_{i}$ in the query, a unigram model is defined as $p(x \vert z) =\prod_{i = 1}^{I}p(w_{i} \vert z)$. Using this formulation, we can then define different models to estimate $p(w \vert z)$.



\paragraph{Term Matching} As a baseline for $p(w \vert z)$, DC define a \emph{term matching} approach that exploits the fact that many queries in our English datasets share vocabulary with target component vocabulary. A smoothed version of this baseline is defined below, where $f(w \vert z)$ is the frequency of matching terms in the target signature, $f(w \vert \mathcal{C})$ is frequency of the term word in the overall documentation collection, and $\lambda$ is a smoothing parameter (for Jelinek-Mercer smoothing): 
\begin{equation*} p(x \vert z) =\prod_{w \in x}(1-\lambda)f(w \vert z)+\lambda f(w \vert \mathcal{C}) \end{equation*}

\paragraph{Translation Model}

In order to account for the co-occurrence between non-matching words and component terms, DC employ a word-based translation model, which models the relation between natural language words $w_{j}$ and individual component terms $u_{j}$. In this paper, we limit ourselves to sequence-based word alignment models \cite{och2003systematic}, which factor in the following manner: 
%
\begin{equation*} p(x \vert z) = \prod_{i=1}^{I} \sum_{j=0}^{J} p_{t}(w_{i} \vert u_{j}) p_{d}(l(j) \vert i,I,J)   \end{equation*}
Here each $p_{t}(w_{i} \vert u_{j})$ defines an (unsmoothed) multinomial distribution over a given component term $u_{j}$ for all words $w_{j}$. The function $p_{d}$ is a distortion parameter, and defines a dependency between the alignment positions and the lengths of both input strings. This function, and the definition of $l(j)$, assumes different forms according to the particular alignment model being used. We consider three different types of alignment models each defined in the following way: 
\begin{equation*}
  p_{d}(l(j) \vert ...)=\left\{
  \begin{array}{@{}ll@{}}
    \frac{1}{J+1}  & {\text{(1)}} \\
    a(j \vert i,I,J) & {\text{(2)}} \\
    a(t(j) \vert i,I,tlen(J)) & {\text{(3)}} \\    
  \end{array}\right.
\end{equation*} 
Models (1-2) are the classic IBM word-alignment models of \newcite{brown1993mathematics}. IBM Model 1, for example, assumes a uniform distribution over all positions, and is the main model investigated in DC. For comparison, we also experiment with IBM Model 2, where each $l(j)$ refers to the string position of $j$ in the component input, and $a(..)$ defines a multinomial distribution such that $\sum_{j=0}^{J} a(j \vert i,I,J) = 1.0$. 

We also define a new tree based alignment model (3) that takes into account the syntax associated with the function representations. Each $l(j)$ is the relative tree position of the alignment point, shown as $t(j)$, and $tlen(J)$ is the length of the tree associated with $z$. This approach assumes a tree representation for each $z$. We generated these trees heuristically by preserving the information that is lost when components are converted to a linear sequence representation. An example structure for PHP is shown in Figure 4, where the red solid line indicates the types of potential errors avoided by this model. 

Learning is done by applying the standard EM training procedure of \newcite{brown1993mathematics}.



\begin{figure}

\centering 
\begin{tikzpicture}[scale=0.74]
\begin{scope}[frontier/.style={distance from root=90pt},anchor=center,scale=.8]
\tikzset{level 1/.style={level distance=36pt}}
\tikzset{level 3+/.style={level distance=10pt}}
\Tree [.{\texttt{\textbf{\footnotesize bool ZipArchive::deleteName(string \$name)}}} 
		[.{\texttt{ZipArchive}$_{0}$}  \node(e1){\textbf{ZipArchive}}; ] 
		[.{\texttt{deleteName}$_{1}$} \node(name){\textbf{delete}}; \node(name2){\textbf{name}}; ]
		[.{\texttt{string \$name}$_{2}$}  \node(e3){\textbf{string}}; 
				\node(e4){\textbf{name}}; 
					]
					[.{\texttt{bool}$_{3}$} \textbf{bool} ]
					 ] 
\end{scope}
\begin{scope}[xshift=-10pt,yshift=-102pt,grow'=up,column sep=-.01cm]
\matrix [matrix of nodes, node distance = 2cm]
{
\node[fill=white] (w1) {\footnotesize Delete}; &  & \node[fill=white] (first) {\footnotesize entry}; & \node[fill=white] (comp) {\footnotesize in}; & \node[fill=white] (w2) {\footnotesize an}; & \node[fill=white] (archive) {\footnotesize archive}; & \node[fill=white] {\footnotesize using}; & \node[fill=white] (pair) {\footnotesize its}; & \node[fill=white] (namestr) {\footnotesize name};   \\
};
\end{scope}
\begin{scope}[dashed]
\draw (name)--(w1);
\draw (name2)--(first);
\draw (e4)--(namestr);
\draw (e1)--(archive);
\end{scope}
\begin{scope}[color=red]
\draw (name2)--(namestr);
\end{scope}
\end{tikzpicture}

{\footnotesize
\begin{tabular}{l l l}
\textbf{X$_{012}$} & $\rightarrow$ & $\big \langle$ \textbf{X$_{\tiny \framebox{01}}$} \textbf{X$_{\tiny \framebox{2}}$}, \textbf{X$_{\tiny \framebox{01}}$} \textbf{X$_{\tiny \framebox{2}}$}  \texttt{bool}    $\big \rangle$ \\
\textbf{X$_{01}$} & $\rightarrow$ & $\big \langle$ \textbf{X$_{\tiny \framebox{1}}$} \text{in an} \textbf{X$_{\tiny \framebox{0}}$}, \textbf{X$_{\tiny \framebox{0}}$} \textbf{X$_{\tiny \framebox{1}}$}  $\big \rangle$ \\
\textbf{X$_1$} & $\rightarrow$ & $\big \langle$ Delete  \textbf{X$_{\tiny \framebox{1}}$},  \texttt{delete} \textbf{X$_{\tiny \framebox{1}}$}  $\big \rangle$ \\
\textbf{X$_1$} & $\rightarrow$ & $\big \langle$ entry,  \texttt{name}  $\big \rangle$ \\
\textbf{X$_0$} & $\rightarrow$ & $\big \langle$ archive,  \texttt{ZipArchive}  $\big \rangle$ \\
\textbf{X$_2$} & $\rightarrow$ & $\big \langle$ using its \textbf{X$_{\tiny \framebox{2}}$},  \textbf{X$_{\tiny \framebox{2}}$}$\big \rangle$ \\
\textbf{X$_2$} & $\rightarrow$ & $\big \langle$ name,  \texttt{string \$name}  $\big \rangle$ \\
\end{tabular}
}

\caption{An example tree structure (above) associated with an input component. Below are Hiero rules \cite{chiang2007hierarchical} extracted from the alignment and tree information. } 
\end{figure}
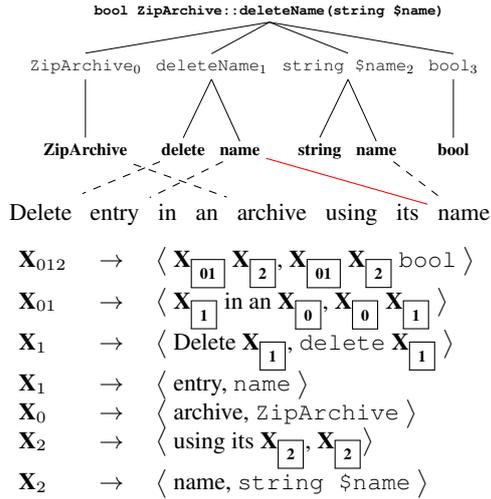


\begin{algorithm}[t]
\algnewcommand\algorithmicinput{\textbf{Input:}}
\algnewcommand\INPUT{\item[\algorithmicinput]}
\algnewcommand\algorithmicoutput{\textbf{Output:}}
\algrenewcommand\algorithmicindent{1.5em}
\algnewcommand\OUTPUT{\item[\algorithmicoutput]}
\caption{Rank Decoder}
\begin{algorithmic}[1]
\INPUT  \text{Query} $x$, Components $\mathcal{C}$ of size $m$, rank $k$, model $\mathcal{A}$, sort function \textsc{K-Best}
\OUTPUT Top $k$ components ranked by $\mathcal{A}$ model score $p$ 
\Procedure{RankComponents}{$x,\mathcal{C},k,\mathcal{A}$}
\State $\textsc{Scores} \leftarrow [\thinspace] $\Comment{Initialize score list}
\For{each component $c$ \Pisymbol{psy}{206} $\mathcal{C}$ }
\State $p \leftarrow \textsc{Align}_{\mathcal{A}}(x,c)$\Comment{Score using $\mathcal{A}$}
\State \textsc{Scores += }$(c,p)$\Comment{Add to list}
\EndFor
\State \textbf{return} K-Best(\textsc{Scores},$k$) \Comment{$k$ best components}
\EndProcedure
\end{algorithmic}
\end{algorithm}

\subsection{Ranking and Decoding}

Algorithm 1 shows how to rank API components. For a text input $x$, we iterate through all known API components $\mathcal{C}$ and assign a score using a model $\mathcal{A}$. We then rank the components by their scores using a \textsc{K-Best} function. This method serves as a type of  word-based decoding algorithm which is simplified by the finite nature of the target language. The complexity of the scoring procedure, lines 3-5, is linear over the number components $m$ in  $\mathcal{C}$. In practice, we implement the \textsc{K-Best} sorting function on line 6 as a binary insertion sort on line 5, resulting in an overall complexity of $O(m \thinspace\thinspace log \thinspace\thinspace m)$. 








While iterating over $m$ API components might not be feasible given more complicated formal languages with recursion, a more clever decoding algorithm could be 
applied, e.g., one based on the lattice decoding approach of \cite{dyer2008generalizing}. Since we are interested in providing initial baseline results, we leave this for future work.

\section{Discriminative Approach}

In this section, we introduce a new model that aims to improve on the previous baseline methods. 

While the previous models are restricted to word-level information, we extend this approach by using a discriminative reranking model that captures phrase information to see if this leads to an improvement. This model can also capture document-level information from the APIs, such as the additional textual descriptions of parameters, \emph{see also} declarations or classes of related functions and syntax information. 

\subsection{Modeling}


Like in most semantic parsing approaches \cite{ZettlemoyerO,liangacl}, our model is defined as a  conditional log-linear model over components $z \in \mathcal{C}$ with parameters $\theta \in \mathbb{R}^{b}$, and a set of feature functions $\phi(x,z)$: $ {p (\thinspace z \vert \thinspace x; \theta) \propto e^{\theta \cdotp \phi \left(x,z\right)} }$. 



Formally,  our training objective is to maximize the conditional log-likelihood of the correct component output $z$ for each input $x$: $\mathcal{O}(\theta)={\sum_{i=1}^{n}}\log\thinspace\thinspace p\left(z_{i}\thinspace|\thinspace x_{i}; \theta \right)$.
%

\subsection{Features}

\begin{figure}
\centering
\begin{tikzpicture}
\matrix(table)[matrix of nodes,
        nodes in empty cells,
        nodes={align=center,text width=1cm},
        row 2/.style={row sep=.1.8em},
        column 1/.style={nodes={text width=.1cm,align=left,font=\small}},
        column 2/.style={nodes={text width=.9cm,align=left,font=\small}},
        column 4/.style={nodes={text width=1.5cm,align=left}},     
        column 5/.style={nodes={text width=1.1cm,align=left,}},
        column 6/.style={nodes={text width=1.8cm,align=left}},
        ampersand replacement=\&,        
    ]{
     \&  \&  \&   \& \&   \\[.1cm]
      {\scriptsize \textbf{z: }} \& \node(p1){\scriptsize \texttt{function}}; \& \node(p2){\scriptsize \texttt{float}}; \& \node(p3){\scriptsize \texttt{\textbf{cosh}}}; \& \node(arg){\scriptsize \texttt{float}}; \& \node(argvar){\scriptsize \texttt{\$\textbf{arg}}}; \\[.4cm]
       {\scriptsize \textbf{x: }} \& \node(1){\scriptsize Returns}; \& \node(w2){\scriptsize the}; \&  \node(w2){\scriptsize \colorbox{gray!10}{\textbf{hyperbolic}}}; \& \node(w3){\scriptsize \colorbox{gray!10}{\textbf{cosine}}}; \& \node(w5){\scriptsize of \colorbox{gray!10}{\textbf{arg}}}; \\
    };
\node(s)[fit=(table-1-4)(table-1-3)]{\tiny \texttt{$c_{4}=$\{\colorbox{blue!15}{\textbf{cosh}},acosh,sinh.\}}};
\node(d)[fit=(table-1-5)(table-1-6)]{\tiny \text{'the \colorbox{blue!15}{arg} of..'}};
\draw(1)--(p1);
\draw[thick]([xshift=6pt,yshift=-8pt]s.north)--([xshift=-12pt,yshift=10pt]p3.south);
\draw[thick]([xshift=6pt,yshift=-8pt]d.north)--([xshift=-13pt,yshift=10pt]argvar.south);
\draw(w2)--([xshift=10pt]p3.south west);
\draw(w3)--([xshift=15pt]p3.south west);
\draw([xshift=-50pt]w5)--(argvar);
\draw[dashed] ([xshift=10pt]w5)--([xshift=-50pt]arg);
\draw[dashed] (1)--(p2);
\draw[dashed] (w2)--([xshift=10pt]p3.south west);
\end{tikzpicture} 
\\ 
\vspace{-.4cm}
{\tiny $\phi$(x,z) = }
 \begin{tabular}{| p{1.28cm} l |}
\hline 
{\tiny \textbf{Model score}:} & {\tiny is it in top 5..10?} \\[-.2cm]
{\tiny \textbf{Pairs/Alignments}:} & {\tiny $(\text{hyperbolic, }\texttt{cosh}) = 1,(\text{cosine, }\texttt{cosh}) = 1, ...$ } \\[-.2cm]
{\tiny \textbf{Phrases}:} & {\tiny $(\text{hyperbolic cosine, }\texttt{cosh}) = 1,(\text{of arg, }\texttt{float \$arg}) = ...$} \\[-.2cm]
{\tiny {\textbf{See also: }}} &  {\tiny $(\text{hyperbolic, }c_{4}=\{\texttt{cos,..}\}) = 1,(\text{arg},c_{4}) = 1$, ...} \\[-.2cm]
{\tiny {\textbf{In Descr.: }}} & {\tiny $(\text{arg, },\texttt{\$arg}) = 1, (\text{arg },\texttt{float}) = 0, ...$} \\[-.2cm]
{\tiny \textbf{Trees/Matches }} & {\tiny (hyperbolic, \texttt{cosh}, \textsc{name\_node}) = 1, number of matches= ...} \\[.1cm]
\hline 
\end{tabular}

\caption{Example features used by our rerankers.  }
\end{figure}
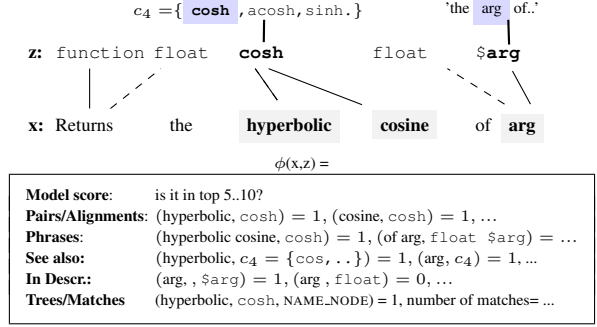

Our model uses word-level features, such as word match, word pairs, as well as information from the underlying aligner model such as Viterbi alignment information and model score. Two additional categories of non-word features are described below. An illustration of the feature extraction procedure is shown in Figure 5 \footnote{A more complete description of features is included as supplementary material, along with all source code.}. 

\paragraph{Phrases Features}

We extract phrase features (e.g., $(\text{hyper. cosine},\texttt{cosh})$ in Figure 5) from example text component pairs by training symmetric word aligners and applying standard word-level heuristics \cite{koehn2003statistical}. Additional features, such as phrase match/overlap, tree positions of phrases, are defined over the extracted phrases. 

We also extract hierarchical phrases \cite{chiang2007hierarchical} using a variant of the SAMT method of \newcite{zollmann2006syntax} and the component syntax trees. Example rules are shown in Figure 4, where \emph{gaps} (i.e., symbols in square brackets) are filled with smaller phrase-tree alignments. 




\paragraph{Document Level Features}

Document features are of two categories. The first includes additional textual descriptions of parameters, return values, and modules. One class of features is whether certain words under consideration appear in the \texttt{@param} and \texttt{@return} descriptions of the target components. For example, the \emph{arg} token in Figure~5 appears in the textual description of the \texttt{\$arg} parameter elsewhere in the documentation string.

Other features relate to general information about abstract symbol categories, as specified in \emph{see-also} assertions, or \emph{hyper-link} pointers. By exploiting this information, we extract general classes of functions, for example the set of hyperbolic function (e.g., \texttt{sinh}, \texttt{cosh}, shown as $c_{4}$ in Figure 5), and associate these classes with words and phrases (e.g., \emph{hyperbolic} and \emph{hyperbolic cosine}). 


\subsection{Learning} 


To optimize our objective, we use Algorithm 2. We estimate the model parameters $\theta$ using a K-best approximation of the standard stochastic gradient updates (lines 6-7), and a ranker function \textsc{RANK}. We note that while we use the ranker described in Algorithm 1, any suitable ranker or decoding method could be used here. 




\begin{algorithm}[t]
\algnewcommand\algorithmicinput{\textbf{Input:}}
\algnewcommand\INPUT{\item[\algorithmicinput]}
\algnewcommand\algorithmicoutput{\textbf{Output:}}
\algnewcommand\OUTPUT{\item[\algorithmicoutput]}
\algnewcommand{\LineComment}[1]{\State \(\triangleright\) #1}
\algrenewcommand\algorithmicindent{1.1em}
\caption{Online Rank Learner}
\begin{algorithmic}[1]
\INPUT  Dataset $D$, components $\mathcal{C}$, iterations $T$, rank $k$, learning rate $\alpha$, model $\mathcal{A}$, ranker function \textsc{Rank} 
\OUTPUT Weight vector $\theta$ 
\Procedure{LearnReranker}{$D,\mathcal{C},T,k,\alpha,\mathcal{A},\textsc{Rank}$}
\State $\theta \leftarrow 0$\Comment{Initialize}
\For{t  \Pisymbol{psy}{206} $1..T$ }
\For{ pairs $(x_{i},z_{i}) \in D$}
\State $\mathcal{S}=$ \textsc{Rank}$(x_{i},\mathcal{C},k,\mathcal{A})$\Comment{Scored candidates}
\State $\Delta = \phi(x_{i},z_{i}) -E_{s \in \mathcal{S} \sim p(s \vert x_{i}; \theta)}[\phi(x_{i},s)]$
\State $\theta = \theta + \alpha\Delta$\Comment{Update online}
\EndFor
\EndFor
\State \textbf{return} $\theta$
\EndProcedure
\end{algorithmic}
\end{algorithm}


\begin{table*}
\newcolumntype{C}[1]{>{\centering\let\newline\\\arraybackslash\hspace{0pt}}m{#1}}
\scriptsize
\setlength{\tabcolsep}{5pt}
\centering 
\begin{tabular}{p{1cm} C{40pt} C{40pt} C{40pt} C{40pt} C{40pt} C{40pt} C{40pt} C{40pt}} 
\hline\hline 
\textbf{Method} & Java  & PHP$_{en}$ & Python & Haskell & Clojure & Ruby & Elisp & C \\ [0.5ex] 
\hline 
\multicolumn{1}{l}{\tiny BOW Model} &
16.4 63.8 31.8  &
 08.0 40.5 18.1  & 
04.1 33.3 13.6 & 
 05.6 55.6 21.7 & 
 03.0 49.2 16.4  &
07.0 38.0 16.9  &
 09.9 54.6 23.5  &
 08.8 48.8 20.0  
\\
\multicolumn{1}{l}{\tiny Term Match} & 
15.7 41.3 24.8 &
 15.6 37.0 23.1  & 
16.6 41.7 24.8 & 
 15.4 41.8 24.0 & 
 20.7 49.2 30.0  &
23.1 46.9 31.2  &
 29.3 65.4 41.4  &
 13.1 37.5 21.9  
\\
{\tiny IBM M1} & 
\textbf{34.3} \textbf{79.8} \textbf{50.2}  &
 \textbf{35.5} \textbf{70.5} \textbf{47.2}  & 
\textbf{22.7} \textbf{61.0} \textbf{35.8} & 
 \textbf{22.3} \textbf{70.3} \textbf{39.6} & 
 \textbf{29.6} \textbf{69.2} \textbf{41.6}  &
 \textbf{31.4} \textbf{68.5} \textbf{44.2}  &
 \textbf{30.6} \textbf{67.4} \textbf{43.5}  &
 21.8 \textbf{63.7} 34.4  
  \\
{\tiny IBM M2} & 
30.3 77.2 46.5  &
 33.2 67.7 45.0  & 
21.4 58.0 34.4 & 
 13.8 68.2 31.8 & 
 26.5 64.2 38.2  &
27.9 66.0 41.4  &
 28.1 66.1 40.7  &
 \textbf{23.7} 60.9 \textbf{34.6}  
  \\
{\tiny Tree Model}  &
29.3 75.4 45.3  &
 28.0 63.2 39.8  & 
17.5 55.4 30.7 & 
 17.8 65.4 35.2 & 
 23.0 60.3 34.4  &
 27.1 63.3 39.5  &
 26.8 63.2 39.7  &
 18.1 56.2 29.4  
 \\ \cdashline{1-9}
{\tiny M1 Descr.}  & 
33.3 77.0 48.7  &
 34.1 71.1 47.2  & 
22.7 62.3 35.9 & 
 23.9 69.5 40.2 & 
 29.6 69.2 41.6  &
 32.5 70.0 45.5  &
 30.3 73.4 44.7  &
 21.8 62.7 33.9  
\\
{\tiny Reranker}  & 
\textbf{35.3} \textbf{81.5} \textbf{51.4}  &
 \textbf{36.9} \textbf{74.2} \textbf{49.3}  & 
\textbf{25.5} \textbf{66.0} \textbf{38.7} & 
 \textbf{24.7} \textbf{73.9} \textbf{43.0} & 
 \textbf{35.0} \textbf{76.9} \textbf{47.9}  &
 \textbf{35.1} \textbf{72.5} \textbf{48.0}  &
 \textbf{37.6} \textbf{80.5} \textbf{53.3}  &
  \textbf{29.7} \textbf{67.4} \textbf{40.1}  
\\
\hline 
\end{tabular}

\vspace{.2cm} 


\begin{tabular}{p{30.6pt} C{40pt} C{40pt} C{40pt} C{40pt} C{40pt} C{40pt} C{40pt} C{40pt}} 
\hline\hline 
\textbf{Method} & Scheme & PHP$_{fr}$  & PHP$_{es}$ & PHP$_{ja}$ & PHP$_{ru}$ & PHP$_{tr}$ & PHP$_{de}$ & Unix  \\ [0.5ex] 
\hline 
{\tiny BOW Model} &
06.1 58.1 21.4 &
06.1 36.9 16.0  &
 05.9 37.8 15.8  & 
04.7 33.2 13.8 & 
 04.4 43.6 16.6 & 
 05.4 43.4 17.6 &
 04.3 39.2 15.3  &
08.6 49.6 21.0  
\\
{\tiny Term Match} & 
25.5 61.2 37.4 &
04.0 15.8 07.7  &
 02.9 10.4 05.4  & 
02.3 11.2 05.2 & 
 01.0 09.3 03.6 & 
 01.4 08.7 03.6  &
03.8 09.4 06.2  &
15.1 33.8 22.4  
\\
{\tiny IBM M1} & 
\textbf{32.1} \textbf{75.5} \textbf{46.2} &
\textbf{32.1} \textbf{65.1} \textbf{43.5}  &
 \textbf{29.5} \textbf{63.7} \textbf{41.2}  & 
\textbf{23.0} \textbf{58.1} \textbf{34.9} & 
20.3 58.4 \textbf{33.3} & 
 \textbf{25.9} \textbf{61.6} \textbf{38.6}  &
\textbf{22.8} \textbf{62.5} \textbf{36.8}  &
 \textbf{30.2} \textbf{66.9} \textbf{42.2}  
  \\
{\tiny IBM M2} & 
29.5 71.4 43.9 &
30.6 62.2 41.2  &
 26.7 59.8 38.3  & 
22.2 56.1 33.3 & 
 18.5 54.5 30.6 & 
 23.3 57.6 35.8  &
19.8 58.6 33.0  &
 23.0 60.4 36.0  
  \\
{\tiny Tree Model}  &
26.1 71.2 40.3 &
27.9 59.3 38.6  &
 25.9 61.0 37.6  & 
22.6 57.8 34.1 & 
 \textbf{20.6} \textbf{59.0} 32.9 & 
 18,9 55.1 32.0  &
18.5 56.0 30.6  &
 23.0 58.2 34.3  
 \\ \cdashline{1-9}
{\tiny M1 Descr.}  & 
33.1 75.5 47.1 &
31.0 64.8 42.7  &
 28.6 64.9 41.1  & 
25.4 60.4 37.0 & 
 \textbf{21.1} 62.6 34.5 & 
 29.1 62.0 \textbf{41.4}  &
26.7 62.0 38.8  &
 \textbf{34.5} 71.9 47.4  
 \\
{\tiny Reranker}  & 
\textbf{34.6} \textbf{77.5} \textbf{48.9} &
\textbf{32.7} \textbf{66.8} \textbf{44.2}  &
 \textbf{30.6} \textbf{66.3} \textbf{42.6}  & 
\textbf{25.8} \textbf{61.8} \textbf{37.8} & 
 21.1 \textbf{66.8} \textbf{35.9} & 
 \textbf{29.9} \textbf{63.8} 41.2  &
\textbf{28.0} \textbf{65.9} \textbf{40.5}  &
 34.5 \textbf{74.8} \textbf{48.5}  
\\
\hline 
\end{tabular}
\\
\vspace{.3cm}
\begin{tabular}{| c | c | c |}
Accuracy @1 & Accuracy @10 & Mean Reciprocal Rank (MRR) \\
\end{tabular}

\label{table:maindataresults} 
\caption{Test results according to the table below. }
\end{table*}

\section{Experimental Setup}


\subsection{Datasets}

\paragraph{Source code documentation}

Our source code documentation collection consists of the standard library for nine programming languages, which are listed in Figure 3. We also use the translated version of the PHP collection for six additional languages, the details of which are shown in Figure 6. The Java dataset was first used in DC, while we extracted all other datasets for this work. 

The size of the different datasets are detailed in both figures. The number of pairs is the number of single sentences paired with function representations, which constitutes the core part of these datasets. The number of descriptions is the number of additional textual descriptions provided in the overall document, such as descriptions of parameters or return values. 



\begin{figure}[t]
{\footnotesize 
\begin{tabular}{p{.9cm} p{1cm} p{.8cm} p{.8cm} p{.8cm} p{.8cm}}
\textbf{Dataset} & \textbf{\# Pairs} & \textbf{\#Descr.} & \textbf{Symbols} & \textbf{Words} & \textbf{Vocab.} \\ \hline
{\small PHP$_{fr}$}  & {\footnotesize 6,155} & {\footnotesize 14,058} & {\footnotesize 7,922} & {\footnotesize 70,800}  & {\footnotesize 5,904} \tabularnewline

{\small PHP$_{es}$} & {\footnotesize 5,823}  &  {\footnotesize 13,285} &  {\footnotesize 7,571} & {\footnotesize 69,882} &  {\footnotesize 5,790}  \tabularnewline
{\small PHP$_{ja}$} & {\footnotesize 4,903}  &  {\footnotesize 11,251} &  {\footnotesize 6,399} & {\footnotesize 65,565} &  {\footnotesize 3,743}  %
\tabularnewline
{\small PHP$_{ru}$} & {\footnotesize 2,549} & {\footnotesize 6,030} & {\footnotesize 3,340} & {\footnotesize 23,105} & {\footnotesize 4,599} %
\tabularnewline
{\small PHP$_{tr}$} & {\footnotesize 1,822}  &  {\footnotesize 4,414} &  {\footnotesize 2,725} & {\footnotesize 16,033} &  {\footnotesize 3,553}  %
\tabularnewline
{\small PHP$_{de}$} & {\footnotesize 1,538} & {\footnotesize 3,733} &   {\footnotesize 2,417} & {\footnotesize 17,460}  & {\footnotesize 3,209} %
\tabularnewline

\end{tabular}} 
\caption{The non-English PHP datasets. }
\end{figure}

We also quantify the different datasets in terms of unique symbols in the target representations, shown as \emph{Symbols}. All function representations and code sequences are linearized, and in some cases further tokenized, for example, by converting out of camel case or removing underscores. 




\paragraph{Man pages} The collection of man pages is from \newcite{unixman} and includes 921 text-code pairs that span 330 Unix utilities and man pages. Using information from the synopsis and parameter declarations, the target code representations are abstracted by type. The extra descriptions are extracted from parameter descriptions, as shown in the \texttt{DESCRIPTION} section in Figure 1, as well as from the \texttt{NAME} sections of each manual. 




\subsection{Evaluation}

For evaluation, we split our datasets into separate training, validation and test sets. For Java, we reserve 60\% of the data for training and the remaining 40\% for validation (20\%) and testing (20\%). For all other datasets, we use a 70\%-30\% split. From a retrieval perspective, these left out descriptions are meant to mimic unseen queries to our model. After training our models, we evaluate on these held out sets by ranking all known components in each resource using Algorithm 1. A predicted component is counted as correct if it matches \emph{exactly} a gold component. 

Following DC, we report the accuracy of predicting the correct representation at the first position in the ranked list (Accuracy @1) and within the top 10 positions (Accuracy @10). We also report the mean reciprocal rank MRR, or the multiplicative inverse of the rank of the correct answer. 

\paragraph{Baselines} 

For comparison, we trained a bag-of-words classifier (the BoW Model in Table 1). This model uses the occurrence of word-component symbol pairs as binary features, and aims to see if word co-occurrence alone is sufficient to for ranking representations. 

Since our discriminative models use more data than the baseline models, which therefore make the results not directly comparable, we train a more comparable translation model, shown as \emph{M1 Descr.} in Table 1,  by adding the additional textual data (i.e. parameter and return or module descriptions) to the models' parallel training data. 


\section{Results and Discussion}

Test results are shown in Table 1. Among the baseline models, IBM Model 1 outperforms virtually all other models and is in general a strong baseline. Of particular note is the poor performance of the higher-order translation models based on Model 2 and the Tree Model. While Model 2 is known to outperform Model 1 on more conventional translation tasks \cite{och2003systematic}, it appears that such improvements are not reflected in this type of semantic translation context. 

The bag-of-words (BoW) and Term Match baselines are outperformed by all other models. This shows that translation in this context is more complicated than simple word matching. In some cases the term matching baseline is competitive with other models, suggesting that API collections differ in how language descriptions overlap with component names and naming conventions. It is clear, however, that this heuristic only works for English, as shown by results on the non-English PHP datasets in Table 1.

We achieve improvements on many datasets by adding additional data to the translation model (M1 Descr.). We achieve further improvements on all datasets using the discriminative model (Reranker), with most increases in performance occurring at how the top ten items are ranked. This last result suggests that  phrase-level and document-level  features can help to improve the overall ranking and translation, though in some cases the improvement is rather modest. 

Despite the simplicity of our semantic parsing model and decoder, there is still much room for improvement, especially on achieving better Accuracy @1. While one might expect better results when moving from a word-based model to a model that exploits phrase and hierarchical phrase features, the sparsity of the component vocabulary is such that most phrase patterns in the training are not observed in the evaluation. In many benchmark semantic parsing datasets, such sparsity issues do not occur \cite{cimiano2009natural}, suggesting that state-of-the-art methods will have similar problems when applied to our datasets. 

Recent approaches to open-domain semantic parsing have dealt with this problem by using paraphrasing techniques \cite{berant2014semantic} or distant supervision \cite{reddy2014large}. We expect that these methods can be used to improve our models and results, especially given the wide availability of technical documentation, for example, distributed within the Opus project \cite{tiedemann2012parallel}.






\begin{figure*}
\centering
\includegraphics[scale=.34]{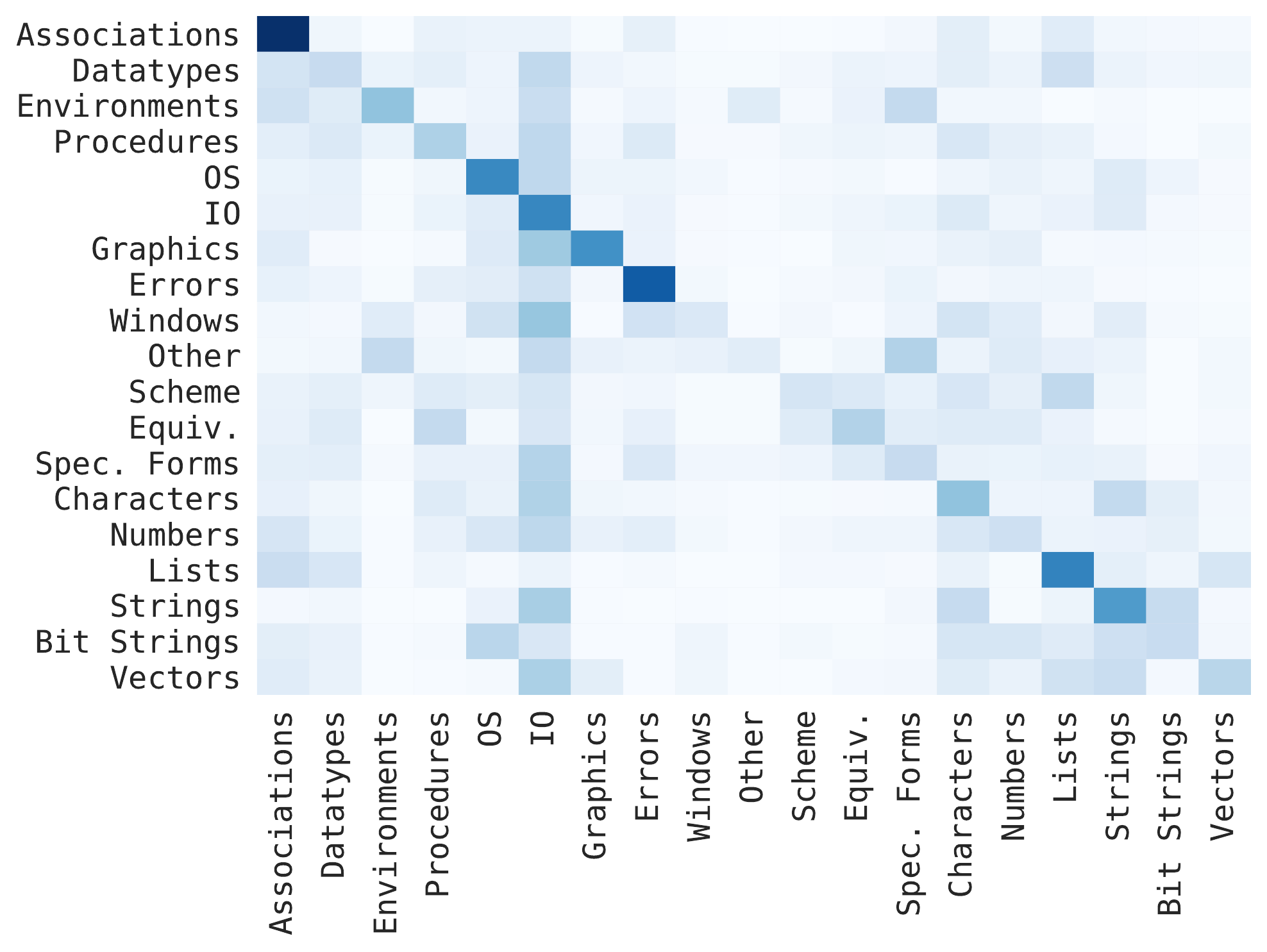}
\includegraphics[scale=.36]{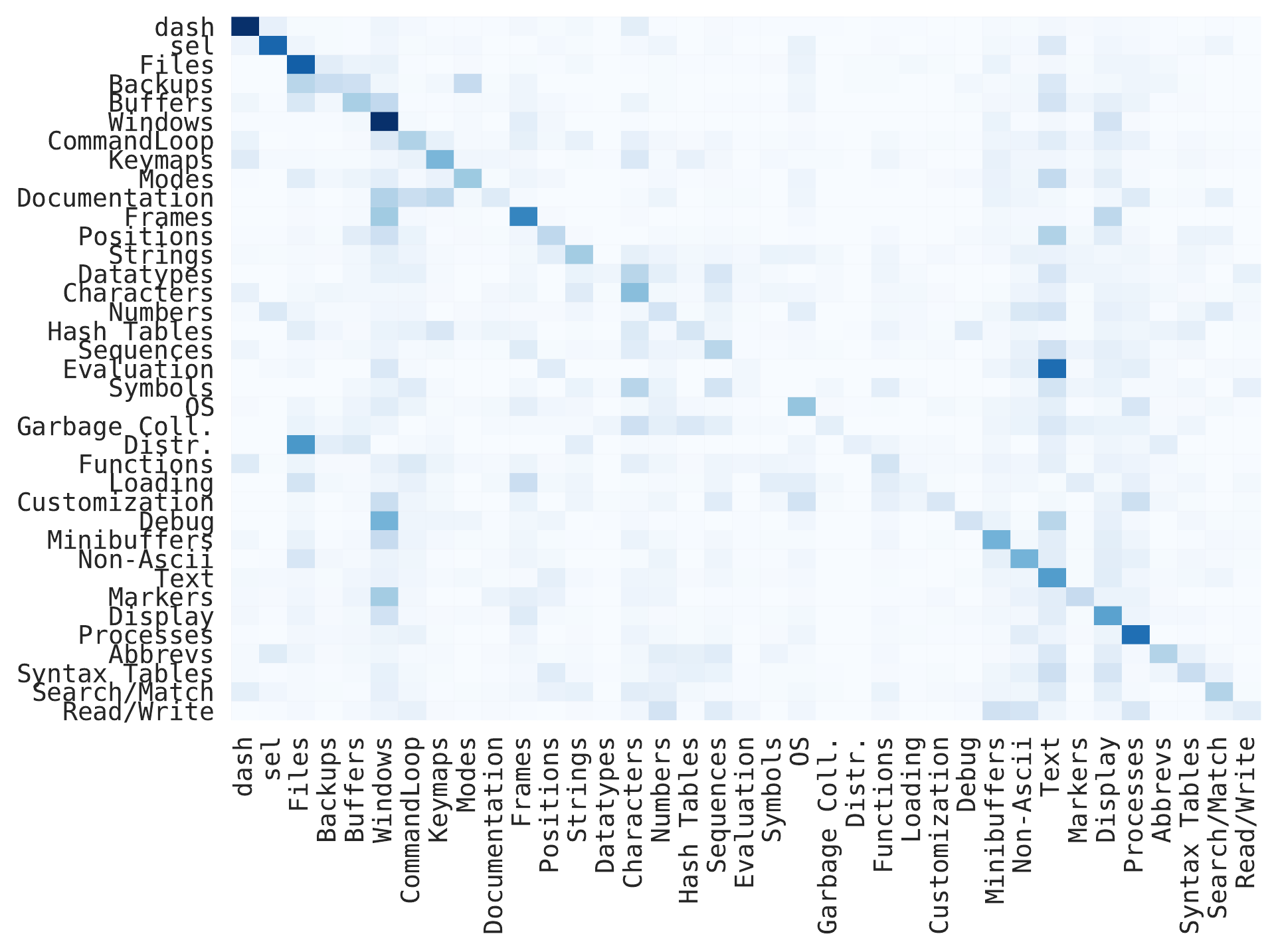} \\ 
\caption{Function predictions by documentation category for Scheme (left) and Elisp (right). }
\end{figure*}

%

\paragraph{Model Errors} We performed analysis on some of the incorrect predictions made by our models. For some documentation sets, such as those in the GNU documentation collection\footnote{https://www.gnu.org/doc/doc.en.html}, information is organized into a small and concrete set of categories/chapters, each corresponding to various features or modules in the language and related functions. Given this information, Figure 7 shows the confusion between predicting different categories of functions, where the rows show the categories of functions to be predicted and the columns show the different categories predicted. We built these plots by finding the categories of the top 50 non-gold (or erroneous) representations generated for each validation example. 


The step-like lines through the diagonal of both plots show that alternative predictions (shaded according to occurrence) are often of the same category, most strikingly for the corner categories. This trend seems stable across other datasets, even among datasets with large numbers of categories. Interestingly, many confusions appear to be between related categories. For example, when making predictions about \texttt{Strings} functions in Scheme, the model often generates function related to \texttt{BitStrings, Characters} and \texttt{IO}. Again, this trend seems to hold for other documentation sets, suggesting that the models are often making semantically sensible decisions. 

Looking at errors in other datasets,  one common error involves generating functions with the same name and/or functionality. In large libraries, different modules sometimes implement that same core functions, such the \texttt{genericpath} or \texttt{posixpath} modules in Python. When generating a representation for the text \emph{return size of file}, our model confuses the \texttt{getsize(filename)} function in one module with others. Similarly, other subtle distinctions that are not explicitly expressed in the text descriptions are not captured, such as the distinction in Haskell between  \emph{safe} and  \emph{unsafe} bit shifting functions. 

While many of these predictions might be correct, our evaluation fails to take into account these various equivalences, which is an issue that should be investigated in future work. Future work will also look systematically at the effect that types (i.e., in statically typed versus dynamic languages) have on prediction. 

\section{Future Work}

We see two possible use cases for this data. First, for benchmarking semantic parsing models on the task of semantic translation. While there has been a trend towards learning executable semantic parsers \cite{berant2013semantic,liang2016learning},  there has also been renewed interest in supervised learning of formal representations in the context of neural semantic parsing models \cite{dong2016language,jia2016data}. We believe that good performance on our datasets should lead to better performance on more conventional semantic parsing tasks, and raise new challenges involving sparsity and multilingual learning.


We also see these resources as useful for investigations into natural language programming. While our experiments look at learning rudimentary translational correspondences between text and code, a next step might be learning to synthesize executable programs via these translations, along the lines of \cite{desai2016program,raza2015compositional}. Other document-level features, such as example input-output pairs, unit tests, might be useful in this endeavor.

\section*{Acknowledgements}

This work was funded by the Deutsche
Forschungsgemeinschaft (DFG) via SFB 732,
project D2. Thanks also to our IMS colleagues,
in particular Christian Scheible, for providing
feedback on earlier drafts, as well as to Jonathan Berant for helpful
discussions.

\bibliography{acl2012}
\bibliographystyle{acl_natbib}

\end{document}